\documentclass{article} 
\usepackage{iclr2025_conference_realign,times}

\usepackage{amsmath,amsfonts,bm}

\def\eqref#1{equation~\ref{#1}}

\def\1{\bm{1}}

\DeclareMathAlphabet{\mathsfit}{\encodingdefault}{\sfdefault}{m}{sl}
\SetMathAlphabet{\mathsfit}{bold}{\encodingdefault}{\sfdefault}{bx}{n}

\usepackage{hyperref}
\usepackage{subcaption}
\usepackage{graphicx}
\usepackage{url}
\usepackage{listings}

\title{The in-context inductive biases of vision-language models differ across modalities}

\author{Kelsey Allen$^{*}$, Ishita Dasgupta$^{*}$, Eliza Kosoy$^{*,1}$, Andrew K. Lampinen\thanks{Equal contribution, ordered alphabetically. $^{1}$Work performed during an internship at Google DeepMind.}\\
Google Deepmind\\
Mountain View, CA, USA\\
\texttt{\{krallen,idg,elko,lampinen\}@google.com}\\ 
}

\iclrfinalcopy 
\begin{document}

\maketitle

\begin{abstract}
Inductive biases are what allow learners to make guesses in the absence of conclusive evidence. These biases have often been studied in cognitive science using concepts or categories -- e.g. by testing how humans generalize a new category from a few examples that leave the category boundary ambiguous. We use these approaches to study generalization in foundation models during in-context learning. Modern foundation models can condition on both vision and text, and differences in how they interpret and learn from these different modalities is an emerging area of study. Here, we study how their generalizations vary by the modality in which stimuli are presented, and the way the stimuli are described in text. We study these biases with three different experimental paradigms, across three different vision-language models. 
We find that the models generally show some bias towards generalizing according to shape over color. This shape bias tends to be amplified when the examples are presented visually. By contrast, when examples are presented in text, the ordering of adjectives affects generalization. However, the extent of these effects vary across models and paradigms.
These results help to reveal how vision-language models represent different types of inputs in context, and may have practical implications for the use of vision-language models.
\end{abstract}

\section{Introduction}

It is impossible for a learner to see every piece of data during training -- they must generalize beyond their experience. Yet the `right' way to generalize is fundamentally under-determined; generalization relies on a system's inductive biases. Many studies show striking differences in how artificial systems and humans generalize \citep[e.g.][]{szegedy2014intriguing,geirhos2019imagenet}, thus demonstrating a mismatch in inductive biases between models and humans, and motivating attempts to bridge this gap \citep{shafahi2019adversarial,geirhos2021partial,muttenthaler2024aligning,fu2024dreamsim}.

One exciting capability of modern foundation models is ``in-context learning'' -- the ability to learn a new concept or task from a few examples (or other cues) presented in context \citep{brown2020language,alayrac2022flamingo,lampinen2024broader}. Few-shot learning involves a particular kind of in-context inductive reasoning. There has been substantial exploration of when models generalize well or poorly from few-shot examples \citep{wei2023larger,zhang2023makes}; these works essentially examine the fit between the inductive biases of the model's learning, the presented examples, and the intended generalization. For example, \citet{chan2022transformers} show that models have different inductive biases for generalizing from information learned in context or in weights. 

Thus, modern vision-language models (VLMs) present an interesting dimension to few-shot learning that has not been thoroughly explored---the effect of the presentation modality and format. If we present examples through images, or as textual descriptions, do these presentation details change how the models generalize? This question is important both practically (understanding how to encourage the generalizations we want) and conceptually (to study differences between the representations of information across modalities).
In this work we therefore study how inductive generalization differs depending on the modality (vision vs. text) in which the data is presented to a vision language model. We also examine the effect of feature order presentation in textual stimuli.

We focus on a well-studied inductive bias: the difference between color and shape features in category learning. Many studies show that humans prefer to generalize along shape rather than dimensions like color or texture, from a very young age \citep{bornstein1985colour,landau1988importance}. That is, in ambiguous situations where either shape or color \emph{could} be the feature determining a category, humans will generally assume that shape is the correct feature to generalize. This tendency in humans has been contrasted with that in convolutional vision models, which tend to prefer lower-level features like texture or color over shape \citep{geirhos2019imagenet,geirhos2020shortcut,hermann2020shapes}.

Recent studies on more modern vision-language models have found that they have recovered some human-like shape bias \citep{gavrikov2024vision}---and some steerability from language. Another recent work \citep{verhoef2024does} found that vision language models do not show the phonological biases in mapping shapes to labels that humans do. However, these studies examine the zero-shot generalization of these models from their training data; not their generalization from categories learned in context. Here, we instead study the shape-color-bias from in-context examples. We also compare to model behavior when the equivalent stimuli are presented described in text rather than directly provided to the model as an image.

We adapt three distinct category learning paradigms (described in detail below) that have been applied in the cognitive literature to study how people represent and generalize concepts. We find that various VLMs show substantial differences in inductive biases across modalities. We find that models show consistent modality biases across different tasks. We also find that the model generalization is affected by the order in which features are mentioned in the text description. However, the exact direction of the biases is idiosyncractic to each model. 

In summary, our contributions are:
\begin{enumerate} \setlength\itemsep{0pt}
    \vspace{-0.7em}
    \item VLMs exhibit different in-context inductive biases when learning from images vs text.
    \item In most cases, models are more shape-biased when learning from images than from text.
    \item Adjective order in text also affects inductive biases: the first descriptor is favored.
    \item However, both the patterns above are overall tendencies which vary in magnitude and even direction across models, task paradigms, and task configuration variables.
    \vspace{-0.7em}
\end{enumerate}

\begin{figure}[t]
    \centering
    \begin{subfigure}{0.5\textwidth}
    \centering
    \includegraphics[width=0.9\linewidth]{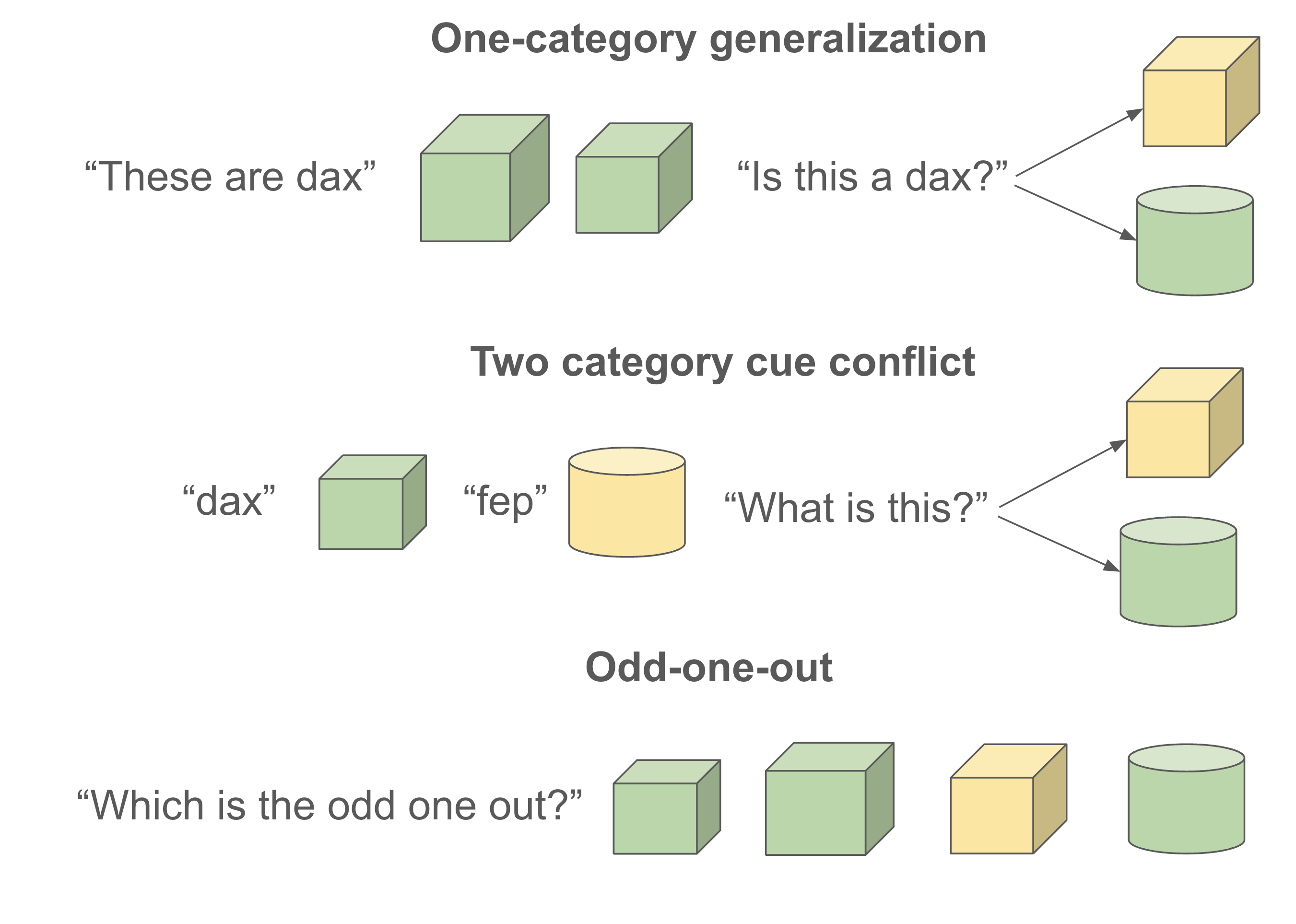}
    \caption{Category-generalization paradigms} \label{fig:overview:paradigms}
    \end{subfigure}%
    \begin{subfigure}{0.5\textwidth}
    \centering
    \includegraphics[width=0.9\linewidth]{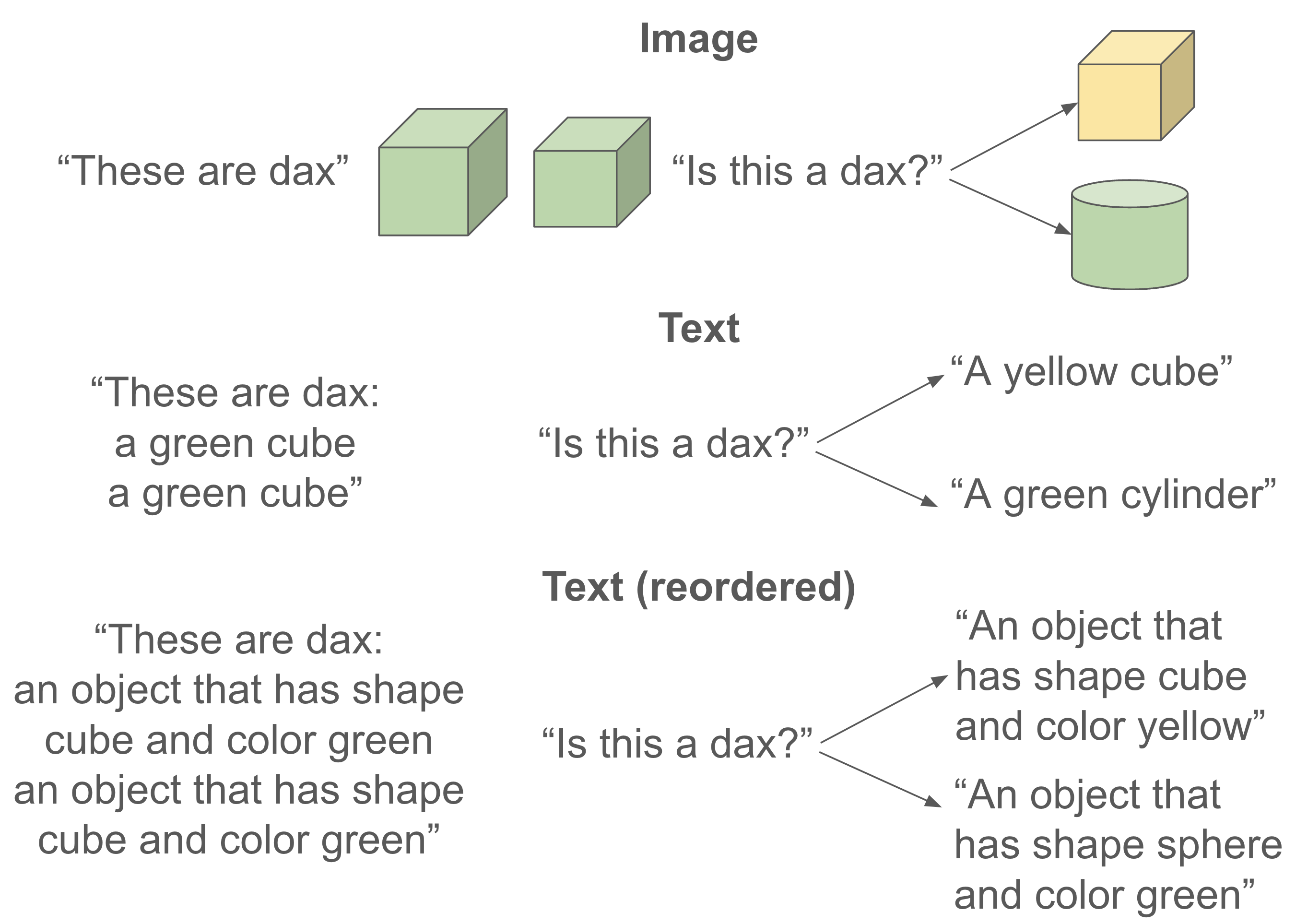}
    \caption{Evaluating different modalities and formats} \label{fig:overview:modalities}
    \end{subfigure}%
    \caption{Conceptual overview of our experimental paradigms.} \label{fig:overview}
    \vspace{-0.5em}
\end{figure}

\section{Methods}

We give an overview of methods here; see Appx. \ref{appx:methods} for details.

\textbf{Image data collection:}
The images were collected from a set of toy geometric shapes. The stimuli in previous shape-bias studies in machine learning models tend to be relatively abstract and difficult to represent in text. To this end, we create a new dataset of simple objects (inspired by category-learning experiments in developmental psychology) with factorial design over color, shape, and angle of photography to enable our study. Creating this dataset from scratch also ensures that the images do not exist in the model pretraining data, and thereby avoids concerns about contamination.

\textbf{Textual descriptions of the objects:}
When we present the objects in text, we use a few phrasing variations. For our primary comparisons we use the standard adjective ordering description, e.g. ``a red cube.'' However, in order to assess whether adjective order affects the model biases, we create descriptions that vary the ordering of adjectives 
of the form ``an object that has color red and shape cube'' or ``an object that has shape cube and color red.'' For the odd-one-out task, we also explore an alternative phrasing ``an object that is red and is cube'' but we observe similar results.

\textbf{Evaluation:}
We prompt the models with the possible answers and an instruction to produce their answer before any explanation, and then evaluate the model outputs by performing a regex search for answers that match the answer pattern.

\textbf{Shape vs. color bias plotting:} We plot our main analyses in terms of the shape-vs-color bias by subtracting the proportion of color generalizations from the proportion of shape generalizations, which makes the comparison clearer across the different task paradigms.

\subsection{Task paradigms}
We consider three main types of tasks (Fig. \ref{fig:overview}): generalizing a single category along different dimensions, generalizing two categories to new instances that have some conflicting features from each of the original categories, and judging which object is the odd-one-out given objects that vary along both features. Note that in every paradigm the correct generalization is ambiguous---there is no direct evidence in the context as to whether shape or color (or their conjunction) is the discriminative feature. Thus, the model generalization in these paradigms offers a measure of which features it prefers to use to form categories in context. Our three paradigms are:

\textbf{One-category generalization:}
We present three images or textual representations of objects with the same color and shape, associated with a nonsense category label. We then evaluate whether the model generalizes this label to objects that match only in shape and objects that match only in color. 

\textbf{Two-category cue conflict:}
We present examples of two different categories that differ in both color and shape; e.g. a green cube as a `dax,' and a yellow sphere as a `fep.' Following \citet{chan2022transformers}, we then test which feature the model uses to generalize to conflicting examples by presenting exemplars that mixes the features of the two categories, e.g. a yellow cube and a green sphere.

\textbf{Odd-one-out:}
We present a set of objects and ask which is the ``odd-one-out''---i.e., the one that does not match the others. We instantiate our odd-one-out tasks to have several objects that are unique in different ways. For example, we might have one object that is a cylinder while the rest are cubes, and another object that is yellow while the rest are green. We can thereby assess whether the model tends to group objects by shape or color by which feature it uses to choose the odd-one-out.

\section{Results} \label{sec:results}
\begin{figure}[th]
\centering
\includegraphics[width=\textwidth]{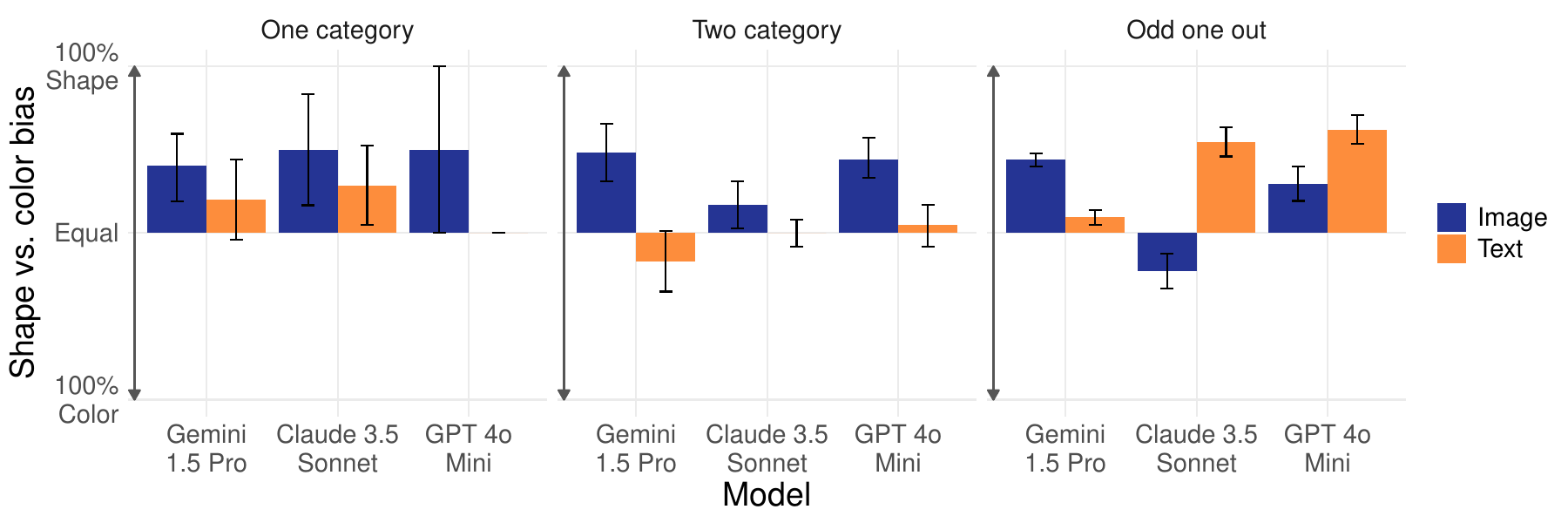}
\caption{Across three category-generalization paradigms, VLMs generalize more by shapes than colors overall (bars above the midline). This bias tends to be amplified when the categories are presented as images (blue), compared to when they are presented in text (orange). However, this pattern is flipped for the odd-one-out task for some models. (Errorbars are bootstrapped 95\%-CIs.)} \label{fig:image_text}
\end{figure}

\begin{figure}[th]
\centering
\includegraphics[width=\textwidth]{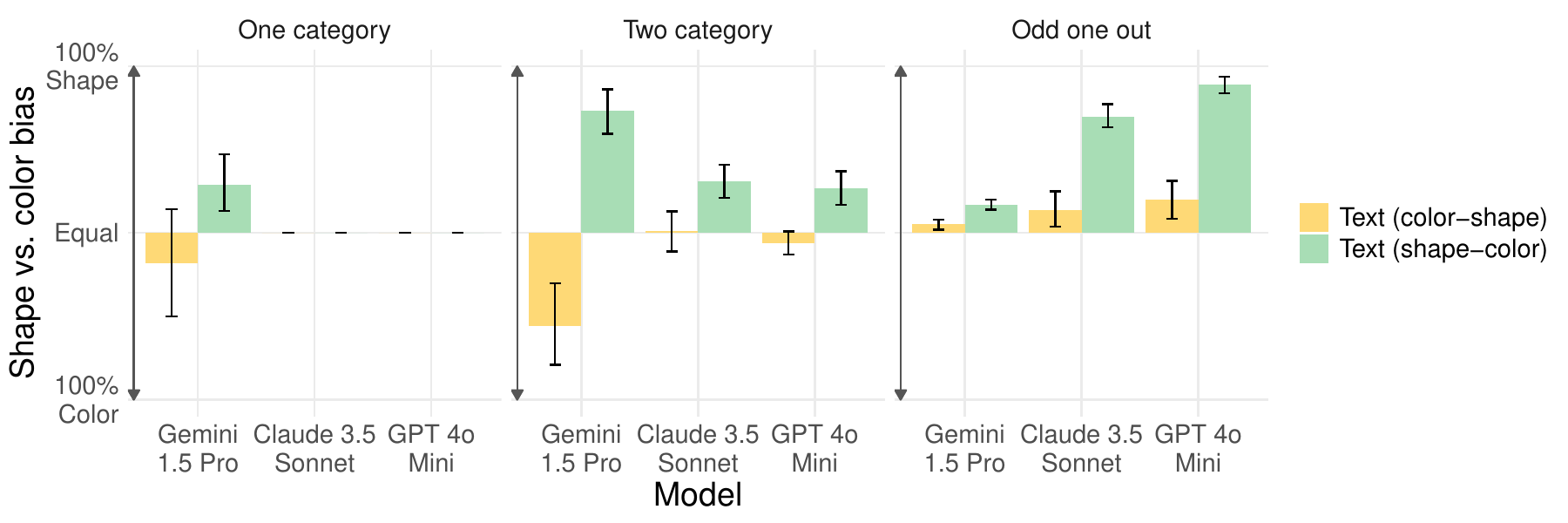}
\caption{The order in which the features are presented in text shifts the VLMs generalization biases; models show some generalization preference toward the feature that is mentioned first (green bars are higher than yellow bars). For one-category tasks, Claude \& GPT refuse all generalizations; hence they show no text-order differences. (Errorbars are boostrapped 95\%-CIs.)} \label{fig:text_order}
\vspace{-0.5em}
\end{figure}

\textbf{Overall differences between images and text:}
In Fig. \ref{fig:image_text} we show an overview of the differences between the inductive biases of the models when learning concepts through image or text modality, across the three different paradigms. The first pattern to note is simply that there is a difference between the modalities; that is, the models' inductive biases seem to differ between the modalities. This overall effect is statistically significant in a regression controlling for task type in all models (respectively; \(t(823) = 12.478, p<0.001\); \(t(823) = 2.432, p=0.015\); \(t(823) = 8.837, p<0.001\)).
The overall pattern is for models to exhibit a stronger shape bias when concepts are presented visually than when they are presented in text (i.e., the blue bars are generally higher on the shape bias axis than the orange bars). This is true for all models for both the explicit category-learning tasks, and for the Gemini model for the odd-one-out task. However, the other two models show a surprising inversion of the pattern for the odd-one-out task, with stronger shape biases from text than images (though the interaction is only statistically significant for Claude, \(t(821) = 2.568, p = 0.010\), not GPT-4o, \(t(821) = 1.17, p = 0.24\)).

\textbf{Feature order in text affects generalization:}
In Fig. \ref{fig:text_order} we show that the order in which features are presented in text has a strong effect on how the models generalize. In general, the models are more shape-biased when shape is mentioned first, and less shape-biased if color is mentioned first (respectively; \(t(1681) = 6.32, p < 0.001\); \(t(839) = 12.1, p<0.001\); \(t(839)=9.23, p<0.001\)). Thus, the models seem to be somewhat biased towards generalizing in accordance with the feature that is mentioned first. However, there is still a weak overall tendency to be shape biased on average.

\textbf{Exploring tasks in more detail:}
In Appx. \ref{appx:detailed_per_task} we present results in more detail for the odd-one-out and one-category tasks (as well as some other variations). We show that there are non-trivial interactions, such as a change in inductive biases depending on set size in the odd-one-out task. 

\section{Discussion}

In this work, we studied the in-context inductive biases of VLMs for concepts presented in images or text. Across three experimental paradigms from the cognitive literature, and three VLMs, we find some commonalities and some idiosyncracies. We find that the models show a moderate shape-over-color bias on average. This shape bias is enhanced when the models are presented with concepts through visual stimuli; when presented with textual stimuli the models are biased by the order in which the features appear in the text. These findings suggest that the way the models represent the input examples is different depending on the format in which they are presented---with the representations from different modalities (or word orders) producing different patterns of generalization.

One possible explanation of these results could relate to the pragmatic interpretation of language (as opposed to vision). In an image, all features are implicitly captured; thus, the presence of a feature does not necessarily indicate that it is meaningful. However, human language (like the text these models are trained) on follows pragmatic conventions, that include an emphasis on stating only relevant and useful information \citep[e.g.][]{grice1975logic}. Indeed, human adjective ordering can be pragmatically determined and interpreted, in some cases with the most discriminative adjectives provided first \citep[e.g.][]{fukumura2018ordering}. 
Thus, the models might be making a pragmatic inference that if the color of an object is mentioned before its shape in describing a category, then color is more important than if shape is mentioned first. It would be interesting to study this hypothesis with a broader set of dimensions and task contexts, as well as relating it to the prevalence of adjectives and their orders in internet text. 
We leave these possible investigations to future work.

Many multimodal models are text first, vision later \citep[e.g.][]{alayrac2022flamingo}; however, all natural systems start with vision and then (in humans) become text-enabled. This fundamental difference posits many interesting questions comparing the two. As foundation models become increasingly capable -- and start to consume information in the myriad ways that humans do -- their similarities and differences to humans will increase, and analyzing them through the lens of cognitive science will become more and more fruitful.

\textbf{Limitations:} These experiments, as a preliminary investigation, have a number of limitations. First, it would be worth evaluating the robustness of the effects across a broader range of tasks, and with other answer formats. Second, it would be worth evaluating a larger range of models, including open-source models. 

\subsection*{Acknowledgements}

We thank Mike Mozer and the reviewers for helpful comments and suggestions.

\bibliography{iclr2025_conference}

\begin{thebibliography}{23}
\providecommand{\natexlab}[1]{#1}
\providecommand{\url}[1]{\texttt{#1}}
\expandafter\ifx\csname urlstyle\endcsname\relax
  \providecommand{\doi}[1]{doi: #1}\else
  \providecommand{\doi}{doi: \begingroup \urlstyle{rm}\Url}\fi

\bibitem[Alayrac et~al.(2022)Alayrac, Donahue, Luc, Miech, Barr, Hasson, Lenc, Mensch, Millican, Reynolds, et~al.]{alayrac2022flamingo}
Jean-Baptiste Alayrac, Jeff Donahue, Pauline Luc, Antoine Miech, Iain Barr, Yana Hasson, Karel Lenc, Arthur Mensch, Katherine Millican, Malcolm Reynolds, et~al.
\newblock Flamingo: a visual language model for few-shot learning.
\newblock \emph{Advances in neural information processing systems}, 35:\penalty0 23716--23736, 2022.

\bibitem[Bornstein(1985)]{bornstein1985colour}
Marc~H Bornstein.
\newblock Colour-name versus shape-name learning in young children.
\newblock \emph{Journal of Child Language}, 12\penalty0 (2):\penalty0 387--393, 1985.

\bibitem[Brown et~al.(2020)Brown, Mann, Ryder, Subbiah, Kaplan, Dhariwal, Neelakantan, Shyam, Sastry, Askell, Agarwal, Herbert{-}Voss, Krueger, Henighan, Child, Ramesh, Ziegler, Wu, Winter, Hesse, Chen, Sigler, Litwin, Gray, Chess, Clark, Berner, McCandlish, Radford, Sutskever, and Amodei]{brown2020language}
Tom~B. Brown, Benjamin Mann, Nick Ryder, Melanie Subbiah, Jared Kaplan, Prafulla Dhariwal, Arvind Neelakantan, Pranav Shyam, Girish Sastry, Amanda Askell, Sandhini Agarwal, Ariel Herbert{-}Voss, Gretchen Krueger, Tom Henighan, Rewon Child, Aditya Ramesh, Daniel~M. Ziegler, Jeffrey Wu, Clemens Winter, Christopher Hesse, Mark Chen, Eric Sigler, Mateusz Litwin, Scott Gray, Benjamin Chess, Jack Clark, Christopher Berner, Sam McCandlish, Alec Radford, Ilya Sutskever, and Dario Amodei.
\newblock Language models are few-shot learners.
\newblock \emph{arXiv preprint arXiv:2005.14165}, 2020.

\bibitem[Chan et~al.(2022)Chan, Dasgupta, Kim, Kumaran, Lampinen, and Hill]{chan2022transformers}
Stephanie~CY Chan, Ishita Dasgupta, Junkyung Kim, Dharshan Kumaran, Andrew~K Lampinen, and Felix Hill.
\newblock Transformers generalize differently from information stored in context vs in weights.
\newblock \emph{MemARI Workshop, NeurIPS 2022}, 2022.

\bibitem[Crutch et~al.(2009)Crutch, Connell, and Warrington]{crutch2009different}
Sebastian~J Crutch, Sarah Connell, and Elizabeth~K Warrington.
\newblock The different representational frameworks underpinning abstract and concrete knowledge: Evidence from odd-one-out judgements.
\newblock \emph{Quarterly Journal of Experimental Psychology}, 62\penalty0 (7):\penalty0 1377--1390, 2009.

\bibitem[Fu et~al.(2024)Fu, Tamir, Sundaram, Chai, Zhang, Dekel, and Isola]{fu2024dreamsim}
Stephanie Fu, Netanel Tamir, Shobhita Sundaram, Lucy Chai, Richard Zhang, Tali Dekel, and Phillip Isola.
\newblock Dreamsim: Learning new dimensions of human visual similarity using synthetic data.
\newblock \emph{Advances in Neural Information Processing Systems}, 36, 2024.

\bibitem[Fukumura(2018)]{fukumura2018ordering}
Kumiko Fukumura.
\newblock Ordering adjectives in referential communication.
\newblock \emph{Journal of Memory and Language}, 101:\penalty0 37--50, 2018.

\bibitem[Gavrikov et~al.(2024)Gavrikov, Lukasik, Jung, Geirhos, Lamm, Mirza, Keuper, and Keuper]{gavrikov2024vision}
Paul Gavrikov, Jovita Lukasik, Steffen Jung, Robert Geirhos, Bianca Lamm, Muhammad~Jehanzeb Mirza, Margret Keuper, and Janis Keuper.
\newblock Are vision language models texture or shape biased and can we steer them?
\newblock \emph{arXiv preprint arXiv:2403.09193}, 2024.

\bibitem[Geirhos et~al.(2019)Geirhos, Rubisch, Michaelis, Bethge, Wichmann, and Brendel]{geirhos2019imagenet}
Robert Geirhos, Patricia Rubisch, Claudio Michaelis, Matthias Bethge, Felix~A Wichmann, and Wieland Brendel.
\newblock Imagenet-trained cnns are biased towards texture; increasing shape bias improves accuracy and robustness.
\newblock In \emph{International Conference on Learning Representations}, 2019.

\bibitem[Geirhos et~al.(2020)Geirhos, Jacobsen, Michaelis, Zemel, Brendel, Bethge, and Wichmann]{geirhos2020shortcut}
Robert Geirhos, J{\"o}rn-Henrik Jacobsen, Claudio Michaelis, Richard Zemel, Wieland Brendel, Matthias Bethge, and Felix~A Wichmann.
\newblock Shortcut learning in deep neural networks.
\newblock \emph{Nature Machine Intelligence}, 2\penalty0 (11):\penalty0 665--673, 2020.

\bibitem[Geirhos et~al.(2021)Geirhos, Narayanappa, Mitzkus, Thieringer, Bethge, Wichmann, and Brendel]{geirhos2021partial}
Robert Geirhos, Kantharaju Narayanappa, Benjamin Mitzkus, Tizian Thieringer, Matthias Bethge, Felix~A Wichmann, and Wieland Brendel.
\newblock Partial success in closing the gap between human and machine vision.
\newblock \emph{Advances in Neural Information Processing Systems}, 34:\penalty0 23885--23899, 2021.

\bibitem[Grice(1975)]{grice1975logic}
HP~Grice.
\newblock Logic and conversation.
\newblock \emph{Syntax and semantics}, 3, 1975.

\bibitem[Hebart et~al.(2020)Hebart, Zheng, Pereira, and Baker]{hebart2020revealing}
Martin~N Hebart, Charles~Y Zheng, Francisco Pereira, and Chris~I Baker.
\newblock Revealing the multidimensional mental representations of natural objects underlying human similarity judgements.
\newblock \emph{Nature human behaviour}, 4\penalty0 (11):\penalty0 1173--1185, 2020.

\bibitem[Hermann \& Lampinen(2020)Hermann and Lampinen]{hermann2020shapes}
Katherine Hermann and Andrew Lampinen.
\newblock What shapes feature representations? exploring datasets, architectures, and training.
\newblock \emph{Advances in Neural Information Processing Systems}, 33:\penalty0 9995--10006, 2020.

\bibitem[Lampinen et~al.(2024)Lampinen, Chan, Singh, and Shanahan]{lampinen2024broader}
Andrew~Kyle Lampinen, Stephanie~CY Chan, Aaditya~K Singh, and Murray Shanahan.
\newblock The broader spectrum of in-context learning.
\newblock \emph{arXiv preprint arXiv:2412.03782}, 2024.

\bibitem[Landau et~al.(1988)Landau, Smith, and Jones]{landau1988importance}
Barbara Landau, Linda~B Smith, and Susan~S Jones.
\newblock The importance of shape in early lexical learning.
\newblock \emph{Cognitive development}, 3\penalty0 (3):\penalty0 299--321, 1988.

\bibitem[Muttenthaler et~al.(2024)Muttenthaler, Greff, Born, Spitzer, Kornblith, Mozer, M{\"u}ller, Unterthiner, and Lampinen]{muttenthaler2024aligning}
Lukas Muttenthaler, Klaus Greff, Frieda Born, Bernhard Spitzer, Simon Kornblith, Michael~C Mozer, Klaus-Robert M{\"u}ller, Thomas Unterthiner, and Andrew~K Lampinen.
\newblock Aligning machine and human visual representations across abstraction levels.
\newblock \emph{arXiv preprint arXiv:2409.06509}, 2024.

\bibitem[Shafahi et~al.(2019)Shafahi, Najibi, Ghiasi, Xu, Dickerson, Studer, Davis, Taylor, and Goldstein]{shafahi2019adversarial}
Ali Shafahi, Mahyar Najibi, Mohammad~Amin Ghiasi, Zheng Xu, John Dickerson, Christoph Studer, Larry~S Davis, Gavin Taylor, and Tom Goldstein.
\newblock Adversarial training for free!
\newblock \emph{Advances in neural information processing systems}, 32, 2019.

\bibitem[Szegedy et~al.(2014)Szegedy, Zaremba, Sutskever, Bruna, Erhan, Goodfellow, and Fergus]{szegedy2014intriguing}
Christian Szegedy, Wojciech Zaremba, Ilya Sutskever, Joan Bruna, Dumitru Erhan, Ian Goodfellow, and Rob Fergus.
\newblock Intriguing properties of neural networks, 2014.
\newblock URL \url{https://arxiv.org/abs/1312.6199}.

\bibitem[Verhoef et~al.(2024)Verhoef, Shahrasbi, and Kouwenhoven]{verhoef2024does}
Tessa Verhoef, Kiana Shahrasbi, and Tom Kouwenhoven.
\newblock What does kiki look like? cross-modal associations between speech sounds and visual shapes in vision-and-language models.
\newblock In \emph{Proceedings of the Workshop on Cognitive Modeling and Computational Linguistics}, pp.\  199--213, 2024.

\bibitem[Wei et~al.(2023)Wei, Wei, Tay, Tran, Webson, Lu, Chen, Liu, Huang, Zhou, et~al.]{wei2023larger}
Jerry Wei, Jason Wei, Yi~Tay, Dustin Tran, Albert Webson, Yifeng Lu, Xinyun Chen, Hanxiao Liu, Da~Huang, Denny Zhou, et~al.
\newblock Larger language models do in-context learning differently.
\newblock \emph{arXiv preprint arXiv:2303.03846}, 2023.

\bibitem[Xu \& Tenenbaum(2007)Xu and Tenenbaum]{xu2007word}
Fei Xu and Joshua~B Tenenbaum.
\newblock Word learning as bayesian inference.
\newblock \emph{Psychological review}, 114\penalty0 (2):\penalty0 245, 2007.

\bibitem[Zhang et~al.(2023)Zhang, Zhou, and Liu]{zhang2023makes}
Yuanhan Zhang, Kaiyang Zhou, and Ziwei Liu.
\newblock What makes good examples for visual in-context learning?
\newblock \emph{Advances in Neural Information Processing Systems}, 36:\penalty0 17773--17794, 2023.

\end{thebibliography}
\bibliographystyle{iclr2025_conference}

\clearpage
\appendix
\section{Supplemental methods} \label{appx:methods}

\subsection{Task paradigms} \label{appx:methods:paradigms}
We describe each paradigm in detail below.

\textbf{One-category generalization:}
Our first task involves a simple kind of underspecified generalization: learning a single category and then testing how models generalize it according to different features. Variations of this paradigm are often used in the cognitive literature \citep[e.g.][]{landau1988importance,xu2007word}. In our instantiation, we present several images or textual representations of objects that have the same color and shape, and associate them with a novel category label. We then evaluate whether the model generalizes this label to objects that match only in shape and objects that match only in color. 

For our main experiments, we use three example instances that all have the same shape and color. In Appx.\ \ref{appx:detailed_per_task} we also present further experiments, inspired by \citet{xu2007word}, on how the model generalizes depending on the variability of features in the exemplars presented in context.

\textbf{Two-category cue conflict:}
In this task, we present examples of two different categories to the model, that differ in both color and shape. For example, we might present a green cube as a `dax,' and a yellow sphere as a `fep.' We then test which feature the model uses to generalize to conflicting examples by presenting exemplars that mix the features of the two categories, e.g. a yellow cube and a green sphere). This follows the inductive generalization paradigm used in \citep{chan2022transformers}. 

\textbf{Odd-one-out:}
Odd-one-out tasks are a longstanding paradigm in cognitive science in which a set of several objects (or images) is provided and participants are asked which is the ``odd-one-out''---i.e., the one that is least like the others \citep[e.g.][]{crutch2009different,hebart2020revealing,muttenthaler2024aligning}. In contrast to the category paradigms above, odd-one-out tasks provide a way of assessing inductive biases without attaching explicit labels to the categories: choosing an object as the odd-one-out effectively implies that the other objects are closer to forming a category. To make an interesting test, we instantiate our odd-one-out tasks to have several objects that are unique in different ways. For example, we might have one object that is a cylinder while the rest are cubes, and another object that is yellow while the rest are green. We can thereby assess whether the model tends to group objects by shape or color by which feature it uses to choose the odd-one-out.

We vary the number of objects in the set \(n\) between 3 and 6---in all cases, there are \(n-2\) objects that have the reference shape and color, one that has the reference shape but a different color (the color odd-one-out), and one that has the reference color but a different shape (the shape odd-one-out).

\subsection{Stimulus images}

\begin{figure}[h!]
    \centering
    \begin{subfigure}{0.25\textwidth}
    \centering
    \includegraphics[width=\linewidth]{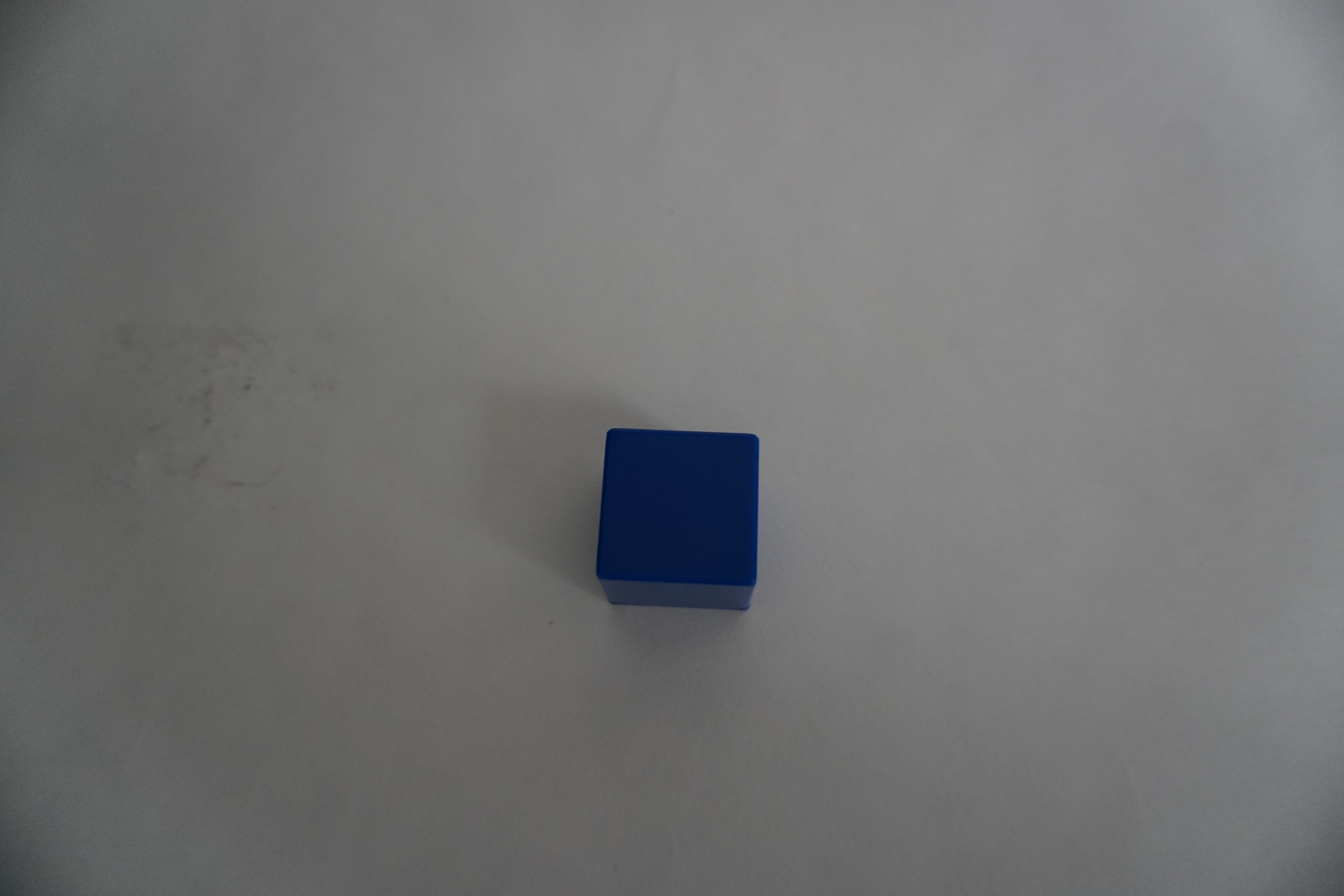}
    \end{subfigure}%
    \begin{subfigure}{0.25\textwidth}
    \centering
    \includegraphics[width=\linewidth]{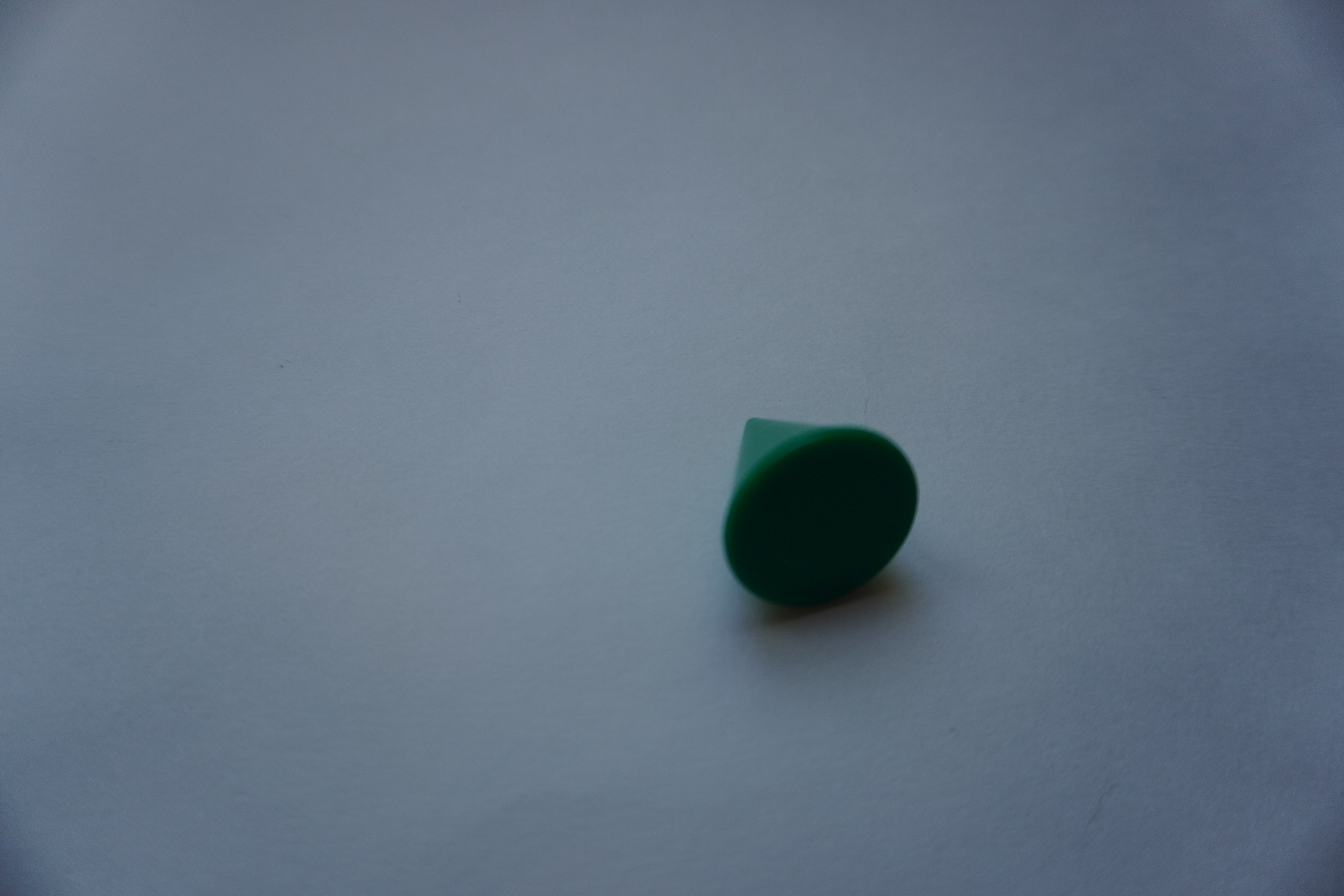}
    \end{subfigure}%
    \begin{subfigure}{0.25\textwidth}
    \centering
    \includegraphics[width=\linewidth]{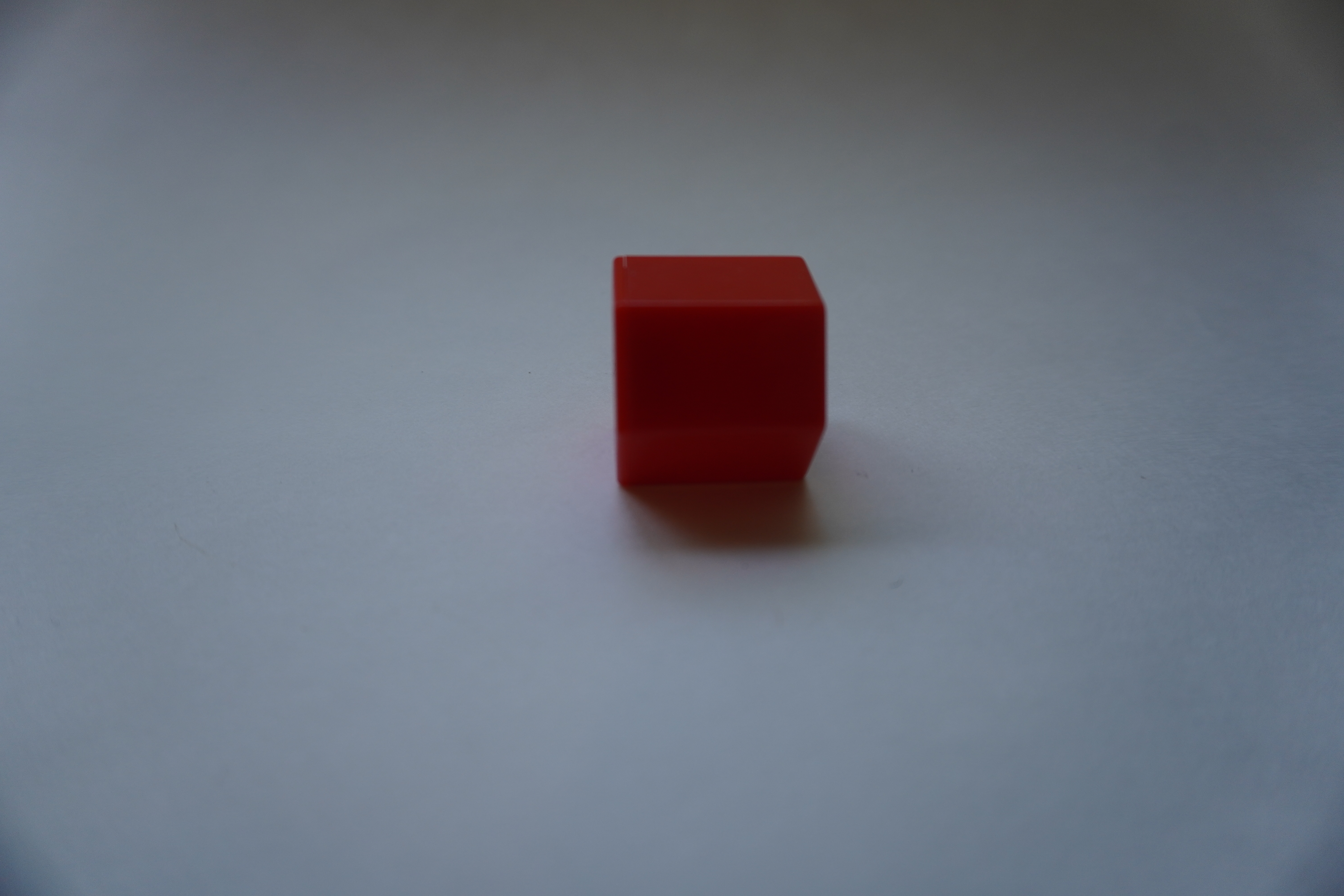}
    \end{subfigure}%
    \begin{subfigure}{0.25\textwidth}
    \centering
    \includegraphics[width=\linewidth]{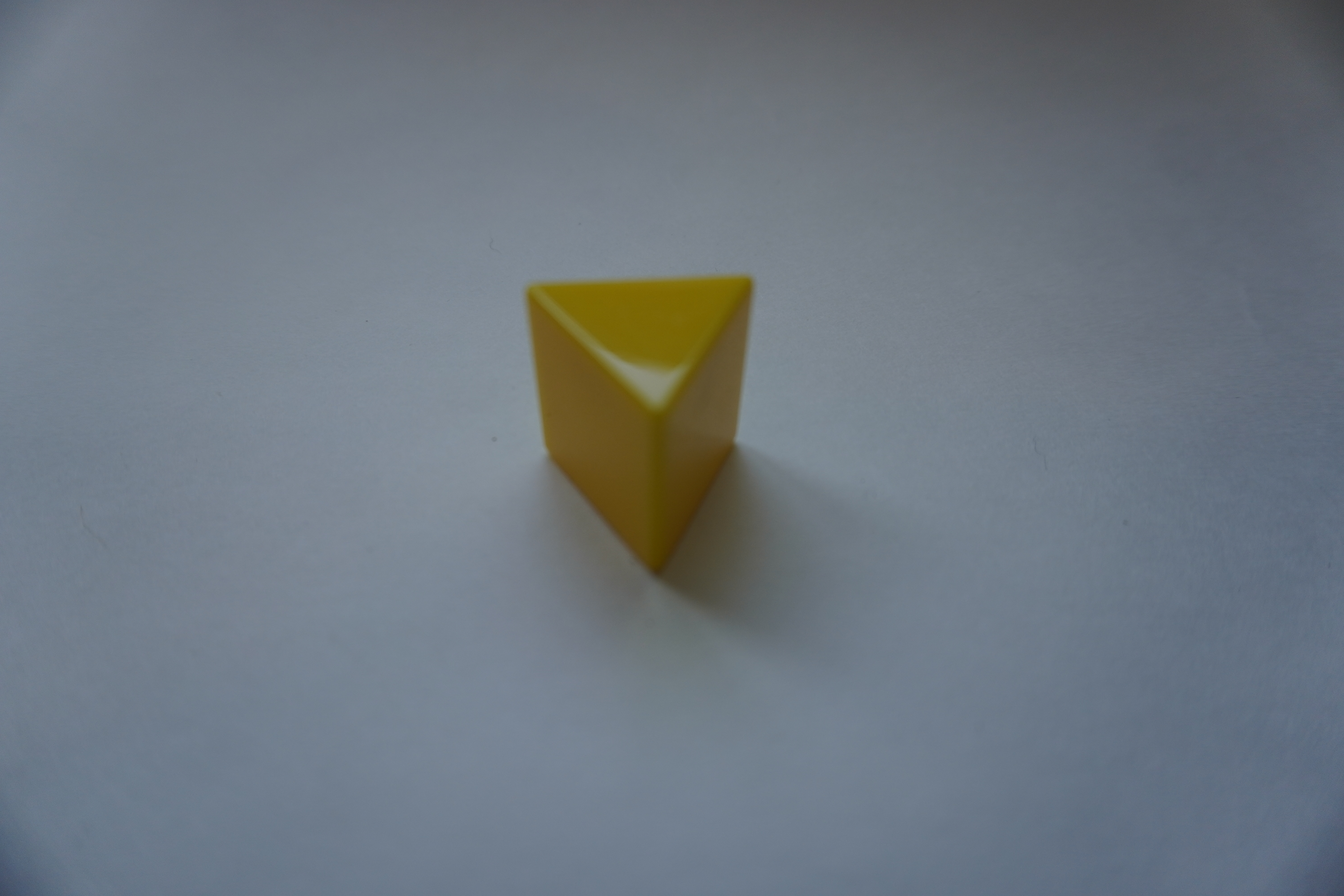}
    \end{subfigure}\\[-0.1em]
    \begin{subfigure}{0.25\textwidth}
    \centering
    \includegraphics[width=\linewidth]{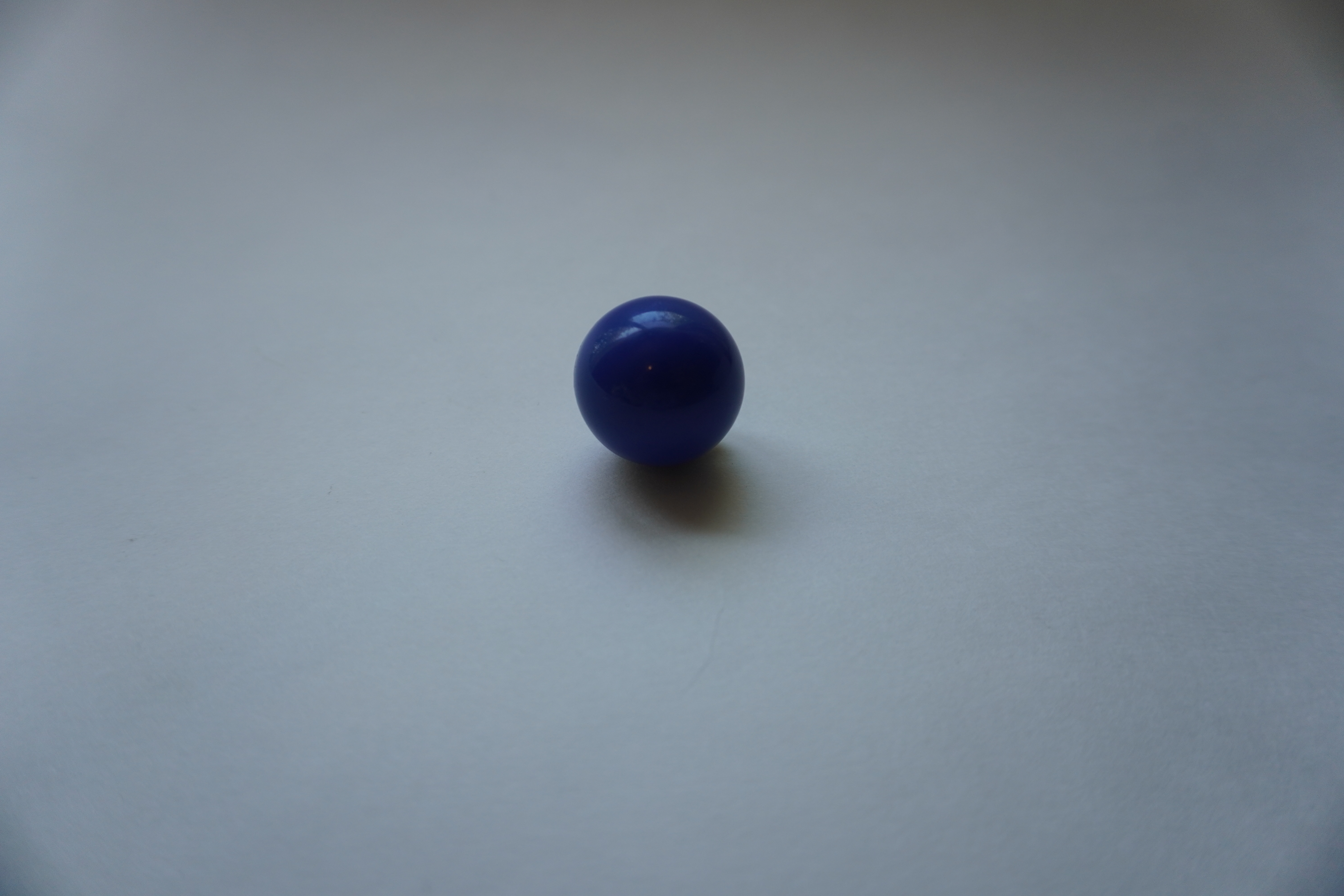}
    \end{subfigure}%
    \begin{subfigure}{0.25\textwidth}
    \centering
    \includegraphics[width=\linewidth]{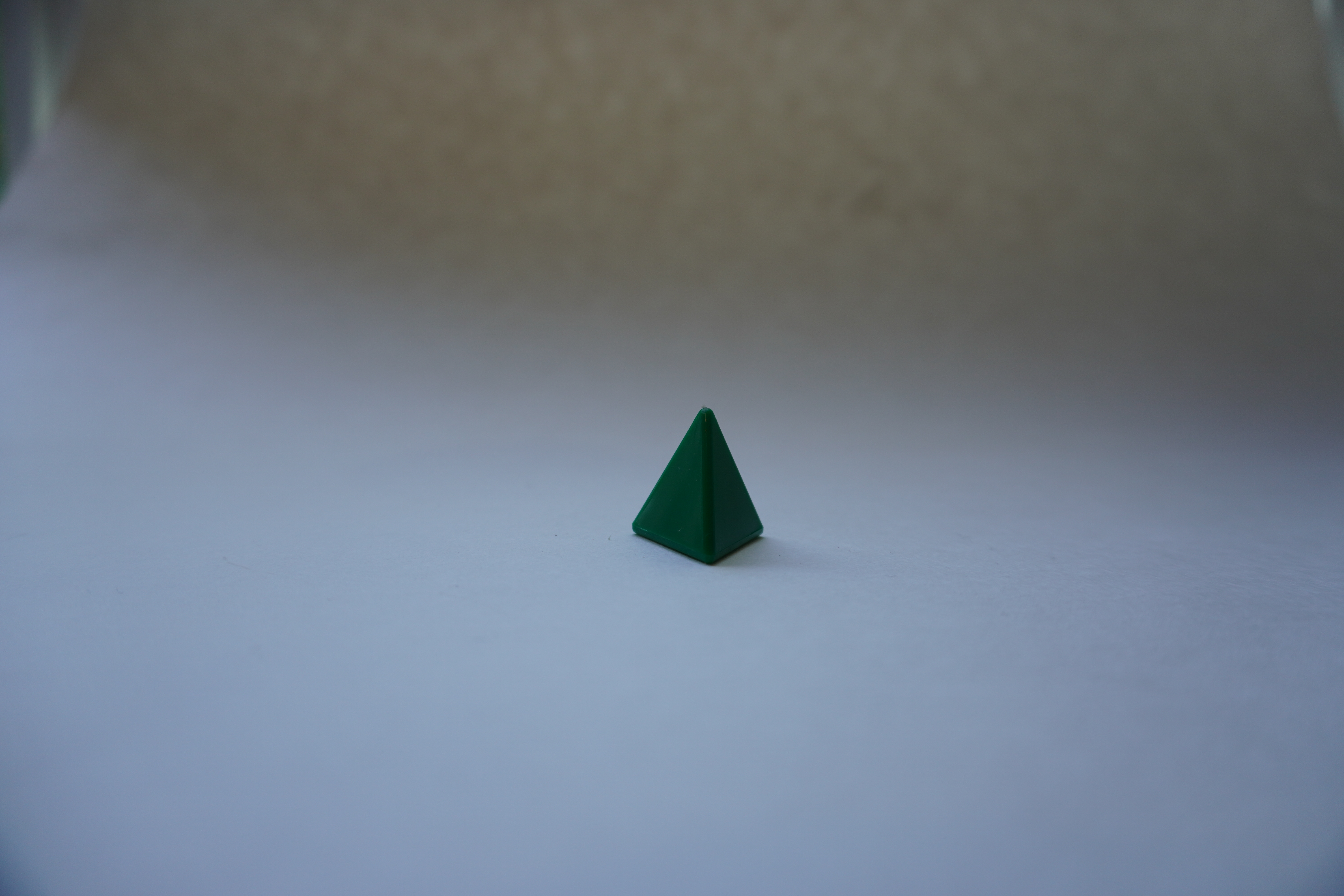}
    \end{subfigure}%
    \begin{subfigure}{0.25\textwidth}
    \centering
    \includegraphics[width=\linewidth]{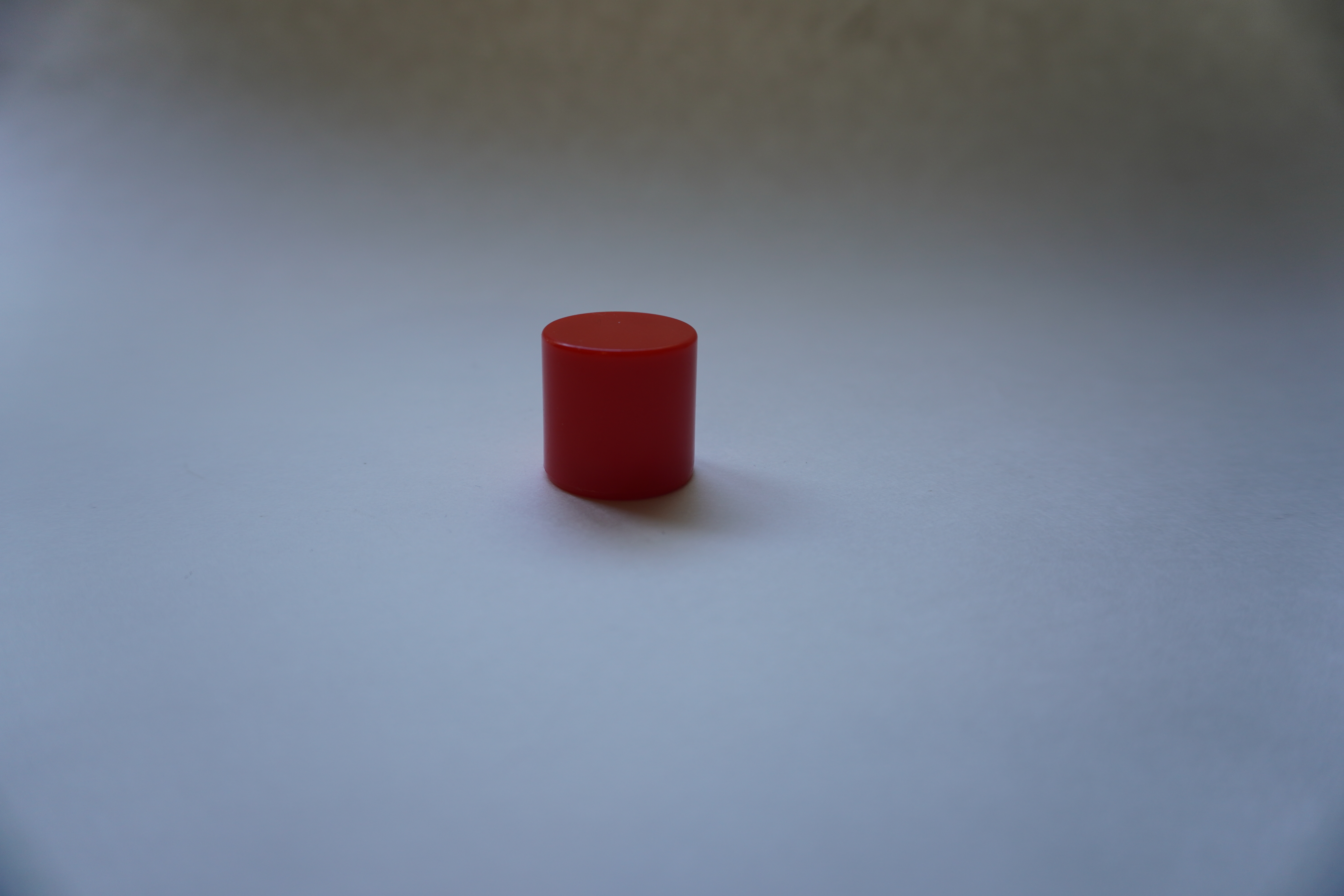}
    \end{subfigure}%
    \begin{subfigure}{0.25\textwidth}
    \centering
    \includegraphics[width=\linewidth]{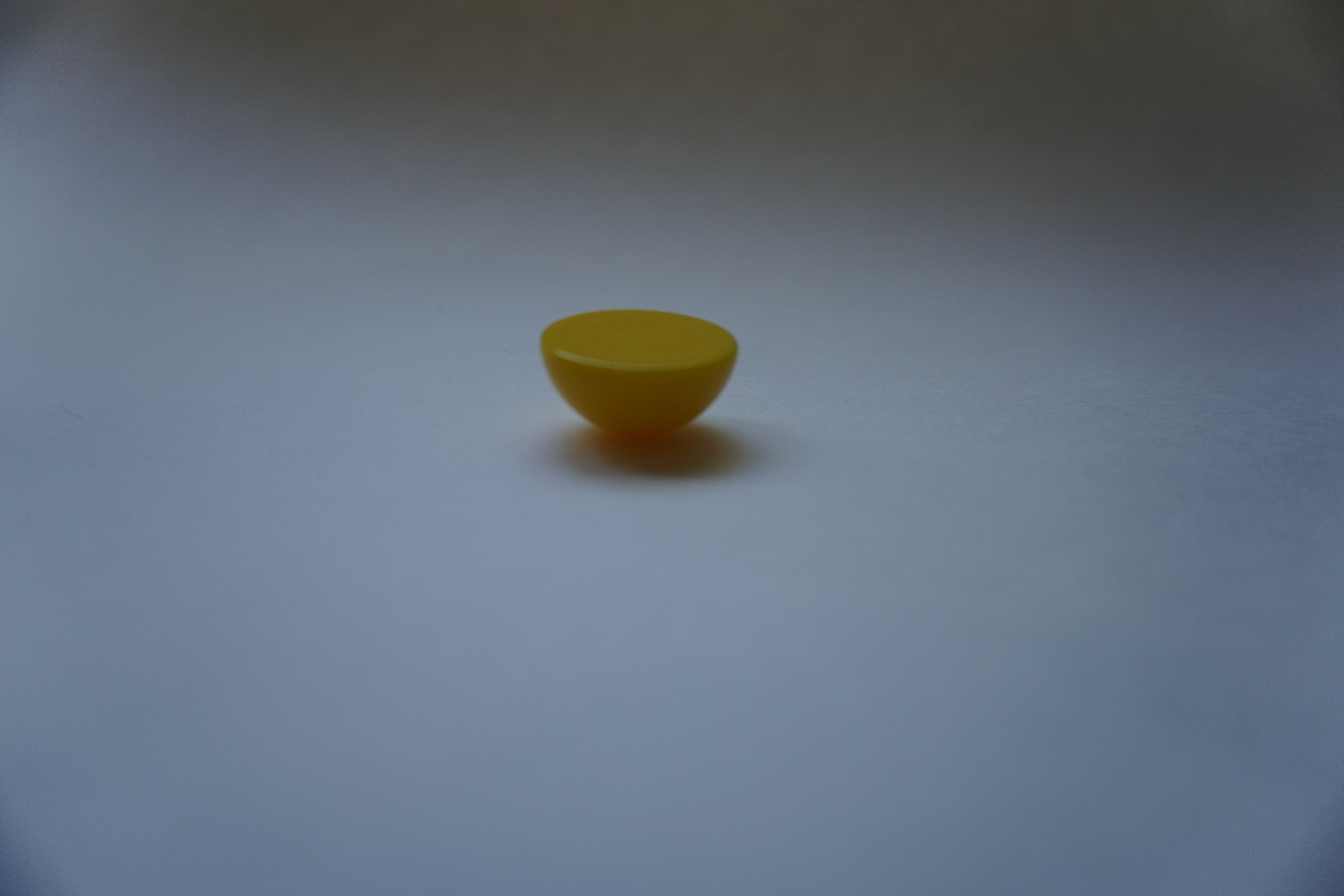}
    \end{subfigure}%
    \caption{Example stimuli used in our experiment, showing some of the range of colors, shapes, and viewpoints presented in the dataset. The variation in viewpoints is intended to ensure that the model is truly recognizing 3D shape rather than relying on canonical orientations.} \label{fig:example_stimuli}
\end{figure}

The images in the dataset consisted of manually taken images of 10 shapes (cone, cube, cylinder, hemisphere, hexagonal prism, pyramid, rectangular prism, sphere, tetrahedron, and triangular prism) in 4 colors (red, yellow, green, and blue), from various sides and the top, taken on a plain white background. See Fig. \ref{fig:example_stimuli} for some examples.

\subsection{Prompts}

The prompt formats used in the experiments are given below:

\begin{lstlisting}[title=Odd one out,breaklines=true,frame=single]
OOO_PROMPT = ['Which of these objects is the odd one out (that is, the one that does not match the others)? Is the odd one out the first, second, third, fourth, fifth, or sixth?'] + images_or_descriptors_list
\end{lstlisting}

\clearpage
\begin{lstlisting}[title=One category generalization,breaklines=true,frame=single]
SINGLE_CATEGORY_PROMPT = [
    f'We are going to play a game where I give you three objects that belong to a category and you have to guess if a new object is also part of that category.\nThese objects are all {name1}:'
] + images_or_descriptors_list + [
    'This is the new object: '
] + new_image_or_descriptor + [
    f"Is this object{name1}?\nIt's ok if you're not sure, just answer with your best guess.\n",
    "Answer in the following format: 'ANSWER: {YES / NO}. \nEXPLANATION: \\\{X}'\n"
]
\end{lstlisting}

\begin{lstlisting}[title=Two category cue conflict,breaklines=true,frame=single]
TWO_CATEGORY_PROMPT = [
    'We are going to play a game where I give you the names of some of objects and you have to guess the name of a new object.\n'
    image_or_description_1,
    f'This object is {name1}',
    image_or_description_2,
    f'This object is {name2}',
    probe_image,
    "What is this object?\nIt's ok if you're not sure, just answer with your best guess.\nAnswer in the format:'ANSWER: This object is {X}.'"
] 
\end{lstlisting}

\section{Detailed performance on individual tasks and other variations} \label{appx:detailed_per_task}

In this section, we break down the performance on the single-category and odd-one-out tasks in more detail (including additional experiments in other conditions) to show the fuller pattern of model behavior. We also show the full pattern of choices rather than the simplified bias metric used in the main plots. There are some quite complex patterns of interaction. Thus, to avoid extremely cluttered plots, we focus these analyses on one particular model (Gemini 1.5).

\subsubsection{Single category} \label{sec:results:per_task:single_category}

\begin{figure}[th]
\centering
\includegraphics[width=0.7\textwidth]{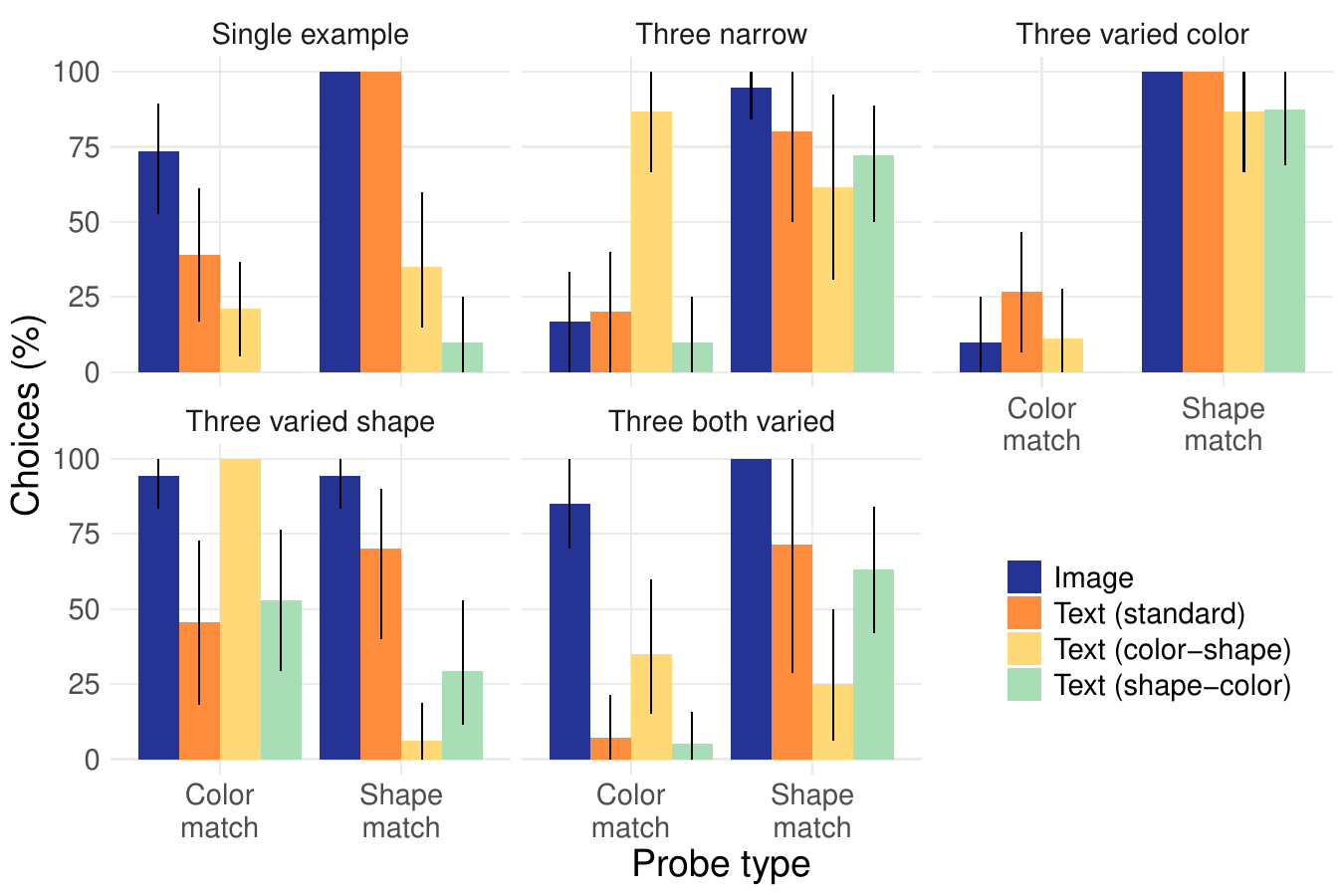}
\caption{Patterns of generalization of a single category presented with varying stimulus sets. There are noticeable changes across stimulus sets, and across modalities, but the patterns are not particularly consistent.} \label{fig:single_cat}
\end{figure}
In this section, we present results on the single category task and several variations. Specifically, a classic paper \citet{xu2007word} has shown that humans learn categories according to the variability in the sampled examples---for example, if all the examples are different types of dogs, the humans will infer that the category is dogs in general, but if all the examples are dalmatians, the humans will infer that the category is dalmatians specifically. We correspondingly tested a range of conditions: providing three narrow examples with the same shape and color (the condition used for the results above), as well four new conditions: only a single example, three examples with varied colors (but consistent shape), three examples with varied shape (but consistent colors), and three examples where both attributes vary (but the probe matches one of the examples along one dimensions).

We show the results in Fig. \ref{fig:single_cat}. The ``color match'' bars show the proportion of the time the model endorses a novel probe stimulus that matches (at least one of) the exemplars on color (but does not match any on shape); the ``shape match'' correspondingly tests for probes that match on shape, but not color. Overall, the patterns of behavior are very noisy and inconsistent, though there is an overall tendency to favor shape matches. However, there is a hint of the expected pattern of generalization when comparing the varied-color and varied shape panels; the model tends to generalize more to matching shapes when the color of examples varies (irrespective of modality), but shows a greater degree of color generalization when the colors are consistent. However, the results when both features vary are surprisingly narrow.

\subsubsection{Odd-one-out cue conflict}

\begin{figure}[ht]
\centering
\includegraphics[width=0.8\textwidth]{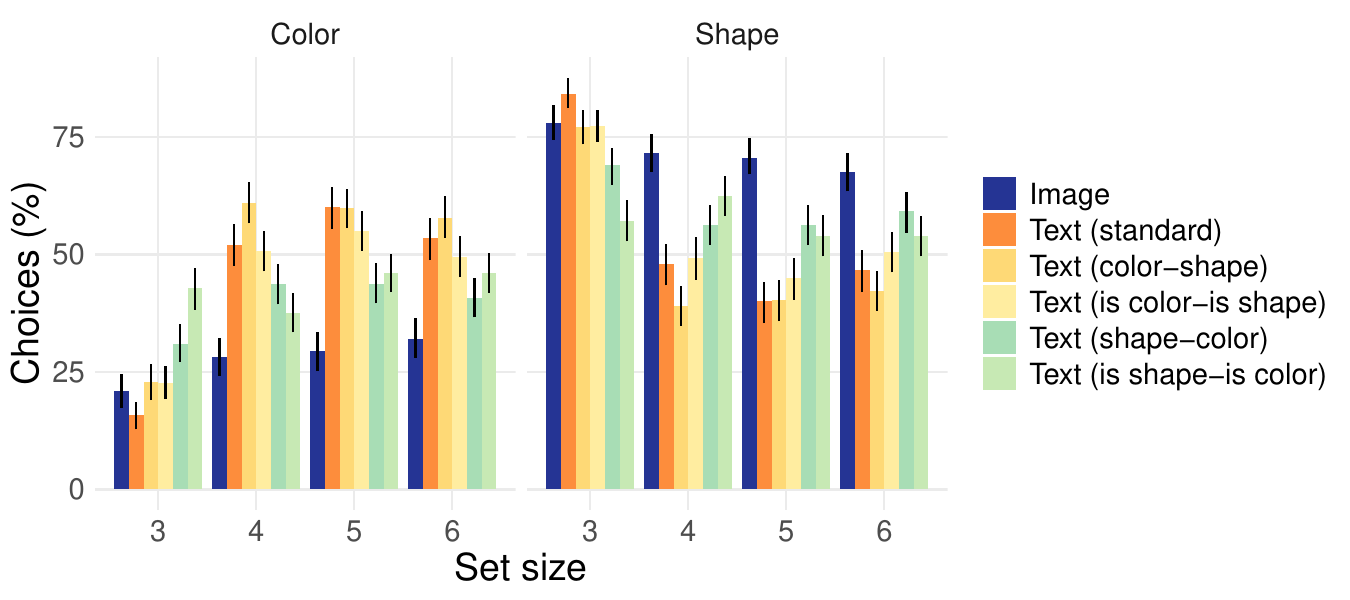}
\caption{Color and shape choices on the Odd-One-Out tasks across set sizes. With four or more objects in the set, the model shows a relatively consistent pattern of biases. Surprisingly, however, when there are three objects in the set, the model shows a stronger shape bias across modalities, and inverted modality patterns overall.} \label{fig:ooo}
\end{figure}

In Fig. \ref{fig:ooo} we show the pattern of results on the odd-one-out tasks. We break the patterns of performance down by set size. While the patterns are quite similar for most set sizes, there is a surprising difference at the smallest possible set size (3); the model shows stronger shape biases overall (across modalities) and the opposite pattern of modality effects. There is no obvious reason why the model would behave differently in this setting, and a qualitative examination of the model outputs does not reveal the answer.

Note that this figure plots raw choices rather than the bias measure in the main text figures. This illustrates that the models are consistently choosing \emph{one of} the unique objects as the odd-one-out. If the models made incorrect choices of objects that did not have any unique feature, the sum of the corresponding bars in the left and right panels would be less than 100\%. However, the models almost always choose either the unique shape or unique-color object as the odd-one-out.

\end{document}